\documentclass[conference]{IEEEtran}
\IEEEoverridecommandlockouts

\usepackage{cite}
\usepackage{amsmath,amssymb,amsfonts}
\usepackage{algorithmic}
\usepackage{graphicx}
\usepackage{textcomp}
\usepackage{xcolor}
\usepackage{array}
\usepackage{float} 
\usepackage{stfloats} 
\usepackage{multirow}
\usepackage{booktabs}
\usepackage{tabularx}
\usepackage{xspace}
\newcommand*{\eg}{e.g.\@\xspace}

\newcommand*{\etal}{et~al.\@\xspace}
\def\BibTeX{{\rm B\kern-.05em{\sc i\kern-.025em b}\kern-.08em
    T\kern-.1667em\lower.7ex\hbox{E}\kern-.125emX}}
\begin{document}

\title{Beyond Stars: Bridging the Gap Between Ratings
and Review Sentiment with LLM
}

\author{
    Najla Zuhir,
    Amna Mohammad Salim,
    Parvathy Premkumar,
    Moshiur Farazi* \\
    \textit{College of Computing \& Information Technology} \\
    \textit{University of Doha for Science and Technology}, Doha, Qatar \\
    \{najla.zuhir, amna.mohammad.salim, parvathypremkumar1901\}@gmail.com, moshiur.farazi@udst.edu.qa
}
\maketitle

\begin{abstract}
We present an advanced approach to mobile app review analysis aimed at addressing limitations inherent in traditional star-rating systems. Star ratings, although intuitive and popular among users, often fail to capture the nuanced feedback present in detailed review texts. Traditional NLP techniques—such as lexicon-based methods and classical machine learning classifiers—struggle to interpret contextual nuances, domain-specific terminology, and subtle linguistic features like sarcasm. To overcome these limitations, we propose a modular framework leveraging large language models (LLMs) enhanced by structured prompting techniques. Our method quantifies discrepancies between numerical ratings and textual sentiment, extracts detailed, feature-level insights, and supports interactive exploration of reviews through retrieval-augmented conversational question answering (RAG-QA). Comprehensive experiments conducted on three diverse datasets (AWARE, Google Play, and Spotify) demonstrate that our LLM-driven approach significantly surpasses baseline methods, yielding improved accuracy, robustness, and actionable insights in challenging and context-rich review scenarios.
\end{abstract}

\begin{IEEEkeywords}
App Reviews, Review Analysis, Discrepancy, Natural Language Processing (NLP), Large Language Models (LLMs), Prompt Engineering, Aspect-Based Sentiment Analysis (ABSA), Topic Modeling, Retrieval-Augmented Question Answering (RAG-QA).
\end{IEEEkeywords}

\section{Introduction}
The exponential growth of mobile applications has transformed app ratings into critical success indicators, directly influencing user adoption decisions and marketplace visibility. Star ratings serve as the primary quality metric for millions of users navigating crowded app stores because they are easy to interpret and provide an immediate signal. However, ratings alone do not provide developers with the detailed feedback required to guide feature prioritization, bug fixes, and overall product strategy for iterative improvement. App store reviews, together with star ratings put the \emph{developer-in-the-loop}. Review texts often reveal specific feature requests, usability issues, and contextual concerns that numeric scores cannot capture. When reviews are analyzed alongside ratings, the combined signal uncovers \emph{what} users think and \emph{why} they think it, which provide contextual information essential for guiding targeted, user-informed improvements.

\begin{figure}[!ht]
  \centering
  \includegraphics[width=\linewidth]{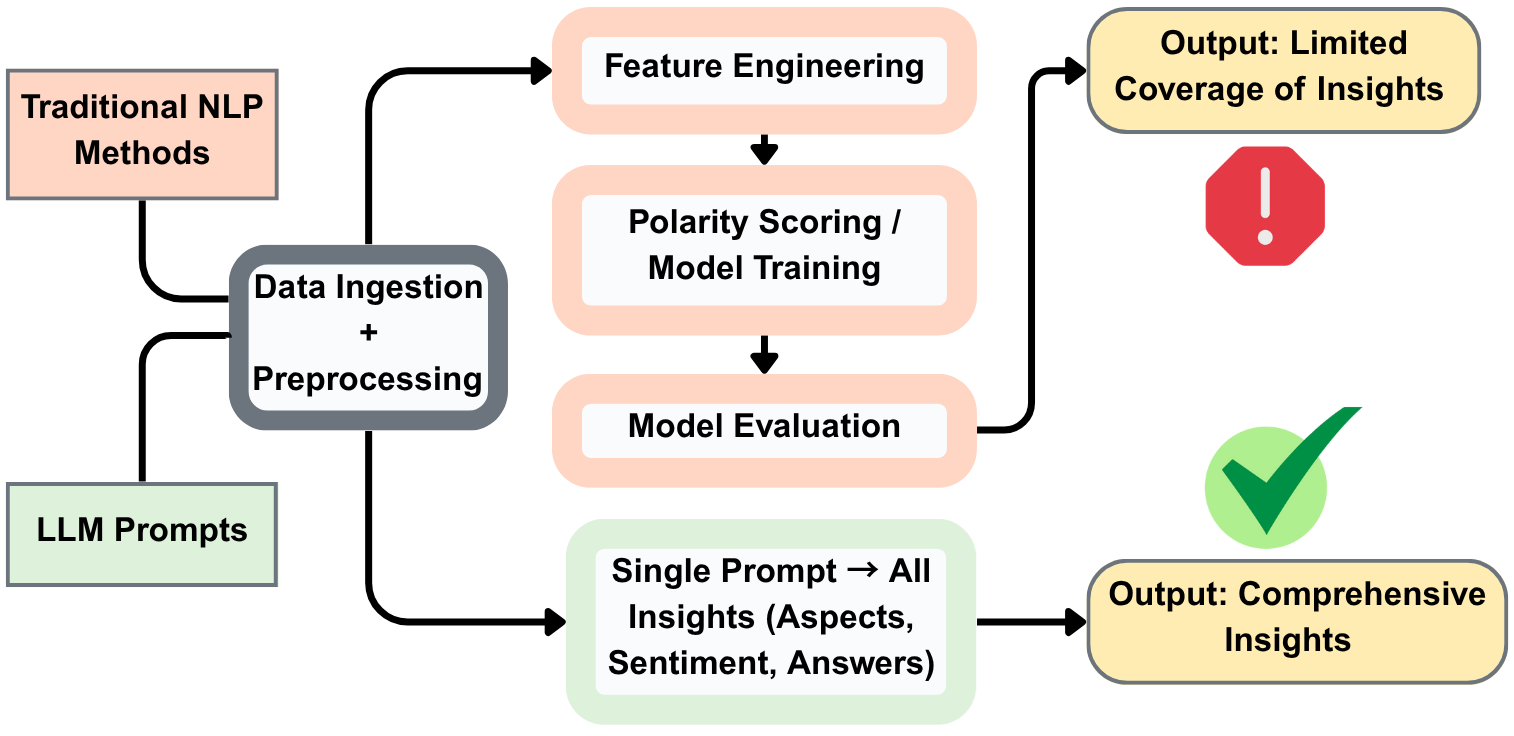}
  \caption{Comparison of traditional NLP pipelines and LLM-based sentiment analysis. Traditional approaches involve multiple sequential steps and require extensive validation, whereas LLM-based methods can perform sentiment extraction in a single, prompt-driven step, yielding more reliable results.}
  \label{fig:ratings‐sentiment}
\end{figure}

Historically, sentiment extraction has depended on lexicon-based methods and conventional machine learning classifiers \cite{biswas2022comparison, kumaresan2023sentiment}. Such techniques perform well on straightforward polarity tasks by identifying positive, negative and/or neutrality of the review text. However, they struggle to capture contextual subtleties, domain-specific terminology, and complex linguistic phenomena such as sarcasm—factors that undermine the reliability of their predictions \cite{yadav2018enhancing}. Furthermore, traditional NLP-based models suffer even more from extracting relevant keywords or information from reviews and communicating effective feedback to the developer. To overcome these shortcomings, recent work has turned to large language models (LLMs) in conjunction with sophisticated prompt-engineering strategies \cite{zhao2024explainability ,agua2025large}. These models exhibit a deeper understanding of context and can distill richer, more actionable insights from unstructured review text.

Structured prompting has emerged as a powerful means of steering LLMs toward fine-grained information extraction and sentiment analysis in app reviews \cite{zhao2024explainability}. By defining explicit prompt templates—ranging from zero-shot instructions (\eg `Classify the sentiment of this review as positive, negative, or neutral.') and few-shot examples to chain-of-thought chains that require the model to articulate its reasoning—these methods can dynamically guide model outputs into precise, structured formats. For example, Wang \etal demonstrated that five-shot prompting of ChatGPT not only rivals but can exceed fine-tuned BERT classifiers on review sentiment tasks. In the aspect-level setting, Shah \etal \cite{shah2024extract} employ an `explain-then-annotate' prompt to extract feature–sentiment pairs directly from unstructured text and show a 23.6 percentage-point F1 gain over rule-based baselines in zero-shot mode and a further ~6 point boost with five examples. Beyond example-based prompting, instructive templates that define the model’s role (e.g. `You are a sentiment analysis assistant…') and enforce output schema (such as JSON) have been shown to improve consistency and ease downstream parsing. Finally, hybrid frameworks that layer domain-specific rules and heuristic filters atop prompt templates enable developers to encode ontological constraints—ensuring that extracted aspects and sentiments align with real-world feature taxonomies and developer needs. Together, these structured prompting strategies unlock rapid, low-overhead extraction of context-sensitive, actionable insights from millions of raw app reviews—without the time and expense of task-specific fine-tuning.

Despite these gains, structured prompting remains vulnerable to brittleness and hallucination across the pipeline. Minor rephrasing can disrupt aspect–sentiment extraction and topic labeling, and generative question-answering (QA) modules may invent spurious features or misinterpret queries. In this work, we address these issues by embedding an automated prompt‐optimization framework that adapts templates to each app’s vocabulary and feedback patterns. At every stage—discrepancy analysis, aspect extraction, topic modeling, and retrieval-augmented QA—we layer in lightweight consistency checks, grounded in VADER‐derived baselines and rule‐based heuristics, to detect and correct unlikely outputs. This hybrid strategy delivers robust, accurate, and scalable review analysis without the need for extensive manual prompt tuning. The contributions of this paper are as follows:
\begin{itemize}
\item \textbf{Baseline discrepancy analysis.} We introduce a lexicon‐based VADER pipeline that computes per‐review polarity, normalizes scores to the five‐star scale, and visualizes the gap between text‐derived sentiment and user ratings—establishing a reproducible baseline for downstream comparison.
\item \textbf{Robust LLM‐driven review mining framework.} We develop a modular, prompt‐based system—leveraging few‐shot examples, prompt chaining, and meta‐prompting—that automatically extracts aspect–sentiment–recommendation triples, uncovers thematic clusters via LLM‐enhanced topic modeling, and supports retrieval-augmented QA. An automated prompt‐optimization loop and lightweight, rule‐based consistency checks ensure stability and accuracy across diverse app vocabularies.
\item \textbf{Comprehensive empirical validation.} Through experiments on three heterogeneous corpora (AWARE, Google Play, and Spotify), we demonstrate that our LLM‐based pipeline significantly outperforms traditional lexicon‐ and machine‐learning baselines in sentiment accuracy, aspect‐extraction F1, topic coherence, and QA relevance.
\end{itemize}

\section{Related Work}
\textbf{Traditional Sentiment Analysis Methods:} Historically, mobile app review analysis primarily utilized lexicon-based sentiment analysis and classical machine learning (ML) techniques, such as Support Vector Machines (SVM) and Random Forest classifiers \cite{biswas2022comparison}. These methods excelled in straightforward polarity classification tasks—categorizing reviews as positive, negative, or neutral—but often struggled with linguistic complexities such as sarcasm, context dependence, and domain-specific jargon \cite{kumaresan2023sentiment,yadav2018enhancing}. Probabilistic topic modeling methods, like Latent Dirichlet Allocation (LDA) and Non-negative Matrix Factorization (NMF), were frequently employed to uncover themes from unlabeled data, but these approaches lacked scalability and robustness against evolving language use \cite{tabianan2022datamining}. To bridge this gap, our work leverages advanced LLM prompting techniques, significantly reducing the dependency on extensive labeled datasets and manual feature engineering, enabling more robust and scalable sentiment analysis.

\textbf{Aspect-Based Sentiment Analysis (ABSA):} ABSA techniques emerged to provide detailed insights by identifying sentiments toward specific app features. Early approaches combined rule-based feature extraction with lexicon-based sentiment scoring, and later methods employed fine-tuned BERT models, substantially improving aspect extraction performance \cite{gao2020emerging,mahmood2020identifying}. Nevertheless, these supervised models required extensive domain-specific annotations, limiting their adaptability and scalability. Recent research has demonstrated the effectiveness of LLMs, notably GPT-4, outperforming traditional ABSA methods without task-specific fine-tuning \cite{shah2024extract}. Our approach extends this advancement by integrating structured LLM prompts that extract not only aspect–sentiment pairs but also actionable user recommendations, delivering richer, developer-focused feedback.

\textbf{Discrepancy Between Star Ratings and Review Text:} Identifying discrepancies between numerical star ratings and textual sentiment remains a critical yet challenging aspect of app review analysis. Approximately 20\% of app reviews exhibit inconsistencies that significantly affect user perception and app marketability \cite{aralikatte2017fault}. Traditional ML methods and advanced deep learning techniques have attempted to address these mismatches but often required extensive preprocessing or lacked nuanced interpretability \cite{ranjan2020comparative,sadiq2021discrepancy}. Recently, the potential of direct discrepancy detection via LLM prompting was suggested but remains largely unexplored \cite{zhao2024explainability}. Our work quantifies rating-review discrepancies by mapping VADER sentiment scores to star ratings and computing absolute differences. This lexicon-based baseline enables future LLM-enhanced analysis.

\textbf{Topic Modeling and Retrieval-Augmented QA:} Beyond traditional sentiment analysis, topic modeling and retrieval-augmented QA methods have seen significant improvements through integration with LLMs. Techniques like BERT-Topic have effectively clustered and summarized large-scale review datasets \cite{agua2025large}. Additionally, retrieval-augmented generative methods have enabled interactive querying of review databases, providing developers immediate insights into user feedback \cite{gao2024retrieval}. Despite these advances, existing methods often lack robustness to minor linguistic variations and suffer from potential hallucinations or inaccuracies in generated responses. Our approach addresses these challenges by embedding automated prompt optimization loops tailored to app-specific vocabularies and implementing consistency checks through rule-based heuristics and lexicon-based baselines, thus ensuring robust and reliable thematic summaries and conversational QA interactions.

\begin{figure*}[!htbp]
  \centering
  
    \includegraphics[width=\linewidth]{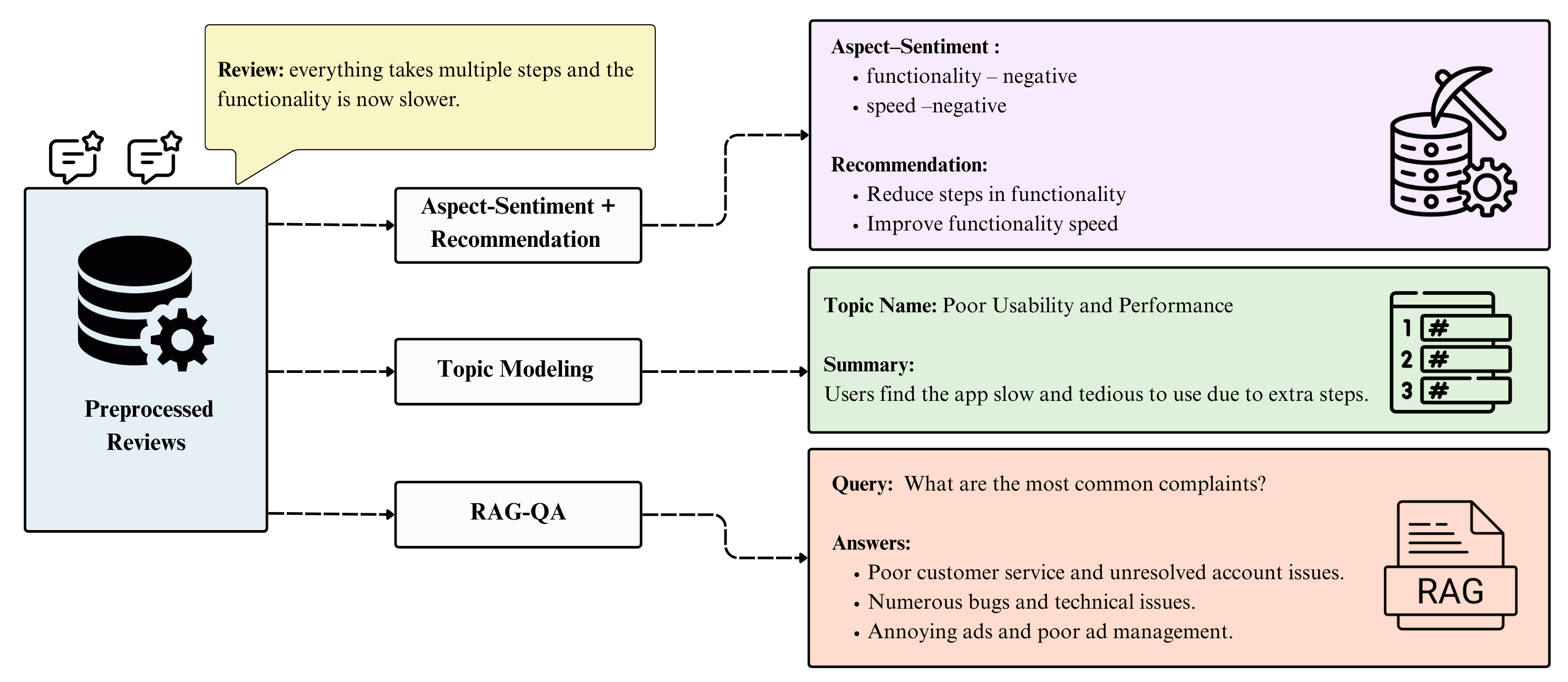}
  \caption{Modular LLM-based review analysis framework. Preprocessed reviews feed into three independent components: (1) Aspect–sentiment extraction with recommendations, (2) topic modeling to surface high-level themes, and (3) retrieval-augmented QA to answer targeted questions.
}
  
  \label{fig:pipeline}
\end{figure*}

\section{Methods}
We first outline the \textit{Aspect Extraction} module for fine-grained feature and sentiment mining. We then detail the \textit{Topic Modeling} approach used to uncover latent thematic structures in large-scale review corpora. Finally, we present the \textit{Retrieval-Augmented QA} system that supports interactive querying and evidence-backed summarization of user feedback.

Although we present the system as a complete LLM-based pipeline for app review analytics, it is intentionally modular: each component can operate independently. For this study, we paired each module with the dataset best suited to its task (AWARE for aspect-level extraction, Spotify for topic modeling and retrieval, and Google Play for discrepancy analysis), ensuring rigorous, task-appropriate evaluation rather than forcing a single dataset through all stages.

\subsection{LLM Backends}
We experimented with the following three LLMs in all our LLM-powered modules: aspect extraction, topic-label generation, and RAG-QA:  
\begin{itemize}
  \item \textbf{GPT-4} (via the OpenAI API), chosen for its state-of-the-art instruction-tuned performance \cite{openai2023gpt4};
  \item \textbf{LLaMA 2 7B-chat} (via HuggingFace), selected as a widely-used open-source alternative \cite{touvron2023llama2};
  \item \textbf{Mistral 7B} (via the official Mistral inference endpoint), included for its strong quality-to-compute trade-off \cite{Jiang2023Mistral7B}.
\end{itemize}

After preliminary evaluations, GPT-4 consistently demonstrated superior performance in capturing contextual nuance and adhering to structured output formats. Therefore, to ensure a clear and focused analysis of the maximum potential of our framework, all experimental results reported in Section V are generated using the GPT-4 model.

\subsection{Aspect Extraction}
Aspect extraction involves identifying specific features and their associated sentiments expressed within app reviews. To achieve accurate and scalable extraction, we employ a structured LLM-based prompting approach. This module takes individual review sentences from the AWARE dataset as input. Using a few-shot prompt that includes exemplar aspect terms drawn directly from AWARE, the model returns structured outputs capturing precise aspect terms. Subsequently, a second LLM prompt classifies the sentiment for each extracted aspect as positive, neutral, or negative. Finally, the module mines explicit user recommendations, outputting imperative phrases for actionable feedback.

\subsection{Topic Modeling}
Topic modeling aims to discover and interpret meaningful thematic clusters within large-scale review corpora. After standard preprocessing steps—cleaning, deduplication, and English-language filtering—the Spotify review dataset serves as input. The processed text undergoes embedding generation using the pre-trained transformer model ("all-mpnet-base-v2") from SentenceTransformers, yielding high-dimensional embeddings \(\mathbf{h}\). BERTopic is applied to cluster these embeddings using UMAP for dimensionality reduction and HDBSCAN for clustering:
\begin{equation}
\mathbf{c} = \text{HDBSCAN}(\text{UMAP}(\mathbf{h})),
\end{equation}
where \(\mathbf{c}\) denotes cluster assignments. To enhance interpretability, the top keywords from each topic cluster are further processed using few-shot prompt, which generates descriptive and intuitive topic labels and summaries, facilitating human comprehension and analysis.

\subsection{Retrieval-Augmented QA}

The retrieval-augmented QA component enables interactive querying of large-scale review datasets. Initially, the cleaned review corpus \(\mathcal{D}\) is segmented into overlapping text chunks, encoded into vector embeddings \(\mathbf{x}_i\) using a sentence transformer model, and indexed within a FAISS-based vector store \(\mathcal{V}\). When presented with a natural-language query \(q\), the retriever computes its embedding \(\mathbf{x}_q\) and retrieves the \(k\) most semantically similar chunks via cosine similarity scoring:
\begin{equation}
\text{score}(q, \mathbf{x}_i) = \frac{\mathbf{x}_q \cdot \mathbf{x}_i}{\|\mathbf{x}_q\| \;\|\mathbf{x}_i\|}, \quad \mathbf{x}_i \in \mathcal{V}.
\end{equation}
These top-ranked chunks populate a structured prompt, refined by meta-prompting techniques, to instruct the answering LLM to produce concise, evidence-backed responses. Thus, users receive targeted, context-aware summaries and actionable insights directly from large datasets without manual review.

\section{Experiments}\label{Experiments}
To evaluate the effectiveness of advanced review analysis techniques, we conducted three LLM-Based experiments across multiple app review datasets. This methodology is designed to address the emerging capabilities of modern LLM-driven approaches. 

Each experimental module is described in detail in the following subsections. 

\subsection{Data Collection \& Pre-processing}
We utilize three heterogeneous app review datasets:

\begin{itemize}
  \item \textbf{Google Play Store App Reviews: } A comprehensive collection of user reviews and corresponding star ratings for various mobile applications. The dataset contains over 12k+ reviews\cite{google_play_reviews_kaggle}.

 \item \textbf{Spotify Reviews: } This large corpus consists of user-generated reviews, each accompanied by metadata such as star rating, date, device, and app version. The Spotify review dataset contains 80k+ reviews\cite{spotify_reviews_kaggle}.

 \item \textbf{AWARE (ABSA Dataset):} human-annotated dataset that contains 11k+ review sentences of apps in the domains of Productivity, Social Networking, and Games. Each sentence is labeled with aspect terms, aspect categories, and aspect sentiment\cite{aware_dataset_zenodo}.
 
\end{itemize}

\textbf{Pre-processing:} Each dataset was processed through a uniform preprocessing pipeline, which included standard text cleaning, deduplication and filtering to retain only English-language reviews. The resulting clean datasets were then used as input for our experiments.
\subsection{Discrepancy Analysis (Rating vs. Sentiment)}

We have experimented with multiple app review datasets, to quantify how often and by what magnitude user reviews diverge from star ratings. For rapid prototyping, we used VADER’s SentimentIntensityAnalyzer \cite{hutto2014vader} to assign a polarity score to each review. These polarity scores were then mapped onto a 1–5 scale (the `Sentiment Rating') to match the original star‐rating range. Finally, we computed a discrepancy feature as the absolute difference between a review sentiment rating and its original star rating, revealing systematic cases where numerical ratings either overestimate or underestimate user satisfaction (Fig. 3). This experiment underscores the importance of combining star ratings with textual analysis: relying solely on numeric scores may misrepresent the true sentiment of users. 

\textbf{Note on the VADER baseline.} We strictly use VADER as a fast, reproducible lexicon baseline. It is known to over-predict neutral in domain-specific app text and to struggle with sarcasm, litotes (e.g., “not bad”), and negation scope; it also lacks coverage for app-specific vocabulary without lexicon adaptation. Consequently, we treat VADER as a conservative floor to motivate discrepancy analysis rather than as an accuracy ceiling.

\begin{figure*}[!ht]
  \centering
  \includegraphics[width=\linewidth]{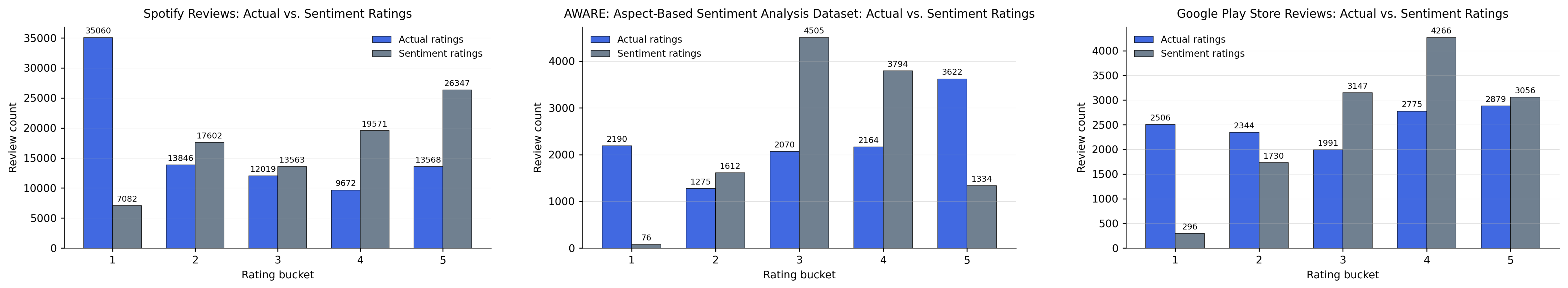}
  \caption{All three datasets shows discrepancy, it suggests that numeric ratings may overestimate or underestimate user satisfaction, as they don’t fully capture the nuanced feedback expressed in written reviews. VADER-derived sentiment ratings, on the other hand, provide a more grounded perspective, highlighting the importance of combining both metrics for an accurate assessment of user satisfaction. }
  \label{fig:ratings-sentiment}
\end{figure*}

\subsection{Aspect Extraction with Recommendation Mining}
Using the AWARE dataset, the Aspect Extraction module was tested through a structured pipeline involving few-shot and chained prompts. Initially, few-shot prompting identified concrete aspect terms within reviews. Subsequently, a chained prompt classified each aspect’s sentiment polarity (positive, neutral, or negative). Finally, recommendation mining extracted actionable feedback expressed by users. The detailed example outputs from this pipeline are presented in Table~\ref{tab:aspect_sent_recomm}. The experimental evaluation aims to measure precision and recall of the extracted (aspect, sentiment) tuples to validate the robustness and utility of our structured prompting methods.

\subsection{LLM-Enhanced Topic Modeling}

In this experiment, our aim was to apply BERTopic for unsupervised topic modeling, enhanced with custom embeddings and post-processed using LLM. The dataset used is spotify reviews, and sentence embeddings are implemented by using a transformer model from the sentence-transformers library (e.g., "all-mpnet-base-v2"). These embeddings are further clustered using HDBSCAN, and topics are extracted via BERTopic. In an effort to improve topic interpretability, a custom function sends the top keywords from each topic to the LLM, using a structured prompt that asks the model to return a short, specific label in title case. Along with this, another LLM prompt—fed with sample reviews from each topic cluster—generates summaries of representative topic documents. The main aim with this integration of BERTopic with LLM-generated labels and summaries is designed to enhance the ease of understanding, readability and user relevance of the discovered topics in the review domain. The detailed example outputs from this pipeline are presented in Table \ref{tab:topic_modelling}. 

\subsection{Retrieval-Augmented QA}

A retrieval-augmented QA system was constructed to enable interactive queries on large-scale review datasets. Reviews were segmented into overlapping text chunks and encoded into vector embeddings indexed in a FAISS-based retriever. Upon receiving a user query, the system employed cosine similarity scoring to retrieve the most relevant text chunks, as described in Eq.~(3). A structured prompt—enhanced via meta-prompting—guided the answering LLM to generate concise, contextually relevant responses referencing the retrieved chunks. Evaluation results are presented in Section V.

\begin{table}[t]
  \centering
  \caption{Sample Retrieval‐Augmented QA Examples}
  \label{tab:qa_examples}
  \footnotesize
  \setlength{\tabcolsep}{4pt}
  \begin{tabular}{@{}p{0.45\columnwidth} p{0.45\columnwidth}@{}}
    \toprule
    \textbf{Query \& Answer} & \textbf{Retrieved Reviews} \\
    \midrule
    \textbf{Q1:} What specific crashes or freezes do users report most often?
    \textbf{A1:} Users report frequent crashes and freezes requiring restarts, often when navigating away, during timeouts, or upon screen‐lock.
    & *Randomly crashes and freezes despite latest version *Regular crashes on Pixel when navigating away *App unresponsive with frequent freezes *Crashes more frequent after updates. \\
    \midrule
    \textbf{Q2:} What are the most common user complaints?
    \textbf{A2:} Users cite poor customer service, unresolved billing issues, bugs, intrusive ads, and weak podcast features.
    & *Horrible customer service—still charging my card  
      *Very poor customer experience; app full of bugs  
      *Constant emails and annoying pop‐ups  
      *Podcasts switch too often; excessive commercials  
      *Unskippable ads and a terrible podcast player \\

    \midrule
    \textbf{Q3:} Give me an overview of all the reviews.
    \textbf{A3:} Despite frustration over bugs and instability, users still appreciate music recommendations, though many consider switching due to technical issues and ads.
    & *Still love the playlist suggestions, but crashes make it unusable  
      *The ads are out of control—feel like I’m paying for promotions  
      *Great discovery features, but app stability needs fixes  
      *Considering switching to another service because of constant bugs \\    
    \bottomrule
  \end{tabular}
\end{table}

\begin{table}[t]
  \centering
  \caption{Sample pipeline output showing the original review sentence, extracted aspect–sentiment pairs, and user recommendations.}
  \label{tab:aspect_sent_recomm}
  \footnotesize
  \setlength{\tabcolsep}{3pt}
  \renewcommand{\arraystretch}{1.1}
  \begin{tabularx}{\columnwidth}{@{} 
      >{\raggedright\arraybackslash}p{0.45\columnwidth} 
      >{\raggedright\arraybackslash}p{0.30\columnwidth}
      >{\raggedright\arraybackslash}p{0.15\columnwidth} @{}}
    \toprule
    \textbf{Sentence} 
      & \textbf{Aspect–Sentiment} 
      & \textbf{Reco.} \\
    \midrule
    don’t get me started on finding old documents, a feature that was said to have improved.
      & \texttt{document finding} – negative 
      & improve document search feature \\
    \addlinespace
    everything takes multiple steps and functionality is now slower.
      & \texttt{functionality} – negative, 
        \texttt{speed} – negative 
      & reduce steps; improve speed \\
    \addlinespace
    i could not turn auto save off, and it was not saving even though i had a stable internet connection.
      & \texttt{auto-save function} – negative, 
        \texttt{stable internet} – neutral 
      & fix auto-save feature \\
    \addlinespace
    i use it for all my classes and it saves me money on notebooks and it’s way easier for organization.
      & \texttt{classes} – positive, 
        \texttt{organization} – positive 
      & — \\
    \addlinespace
    it’s annoying that notability doesn’t offer landscape page when wider view is needed.
      & \texttt{landscape page} – negative 
      & offer landscape view \\
    \addlinespace
    the new evernote home for my desktop is amazing and customizable!
      & \texttt{evernote home} – positive 
      & — \\
    \bottomrule
  \end{tabularx}
\end{table}

\begin{table*}[!t]
  \centering
  \caption{LLM-Generated Topic Modeling Results on the Spotify Reviews Dataset}
  \label{tab:topic_modelling}
  \footnotesize
  \setlength{\tabcolsep}{4pt}        
  \renewcommand{\arraystretch}{1.0}  
  \begin{tabularx}{\textwidth}{ 
      >{\raggedright\arraybackslash}p{0.05\textwidth}  
      >{\raggedright\arraybackslash}p{0.07\textwidth}  
      >{\raggedright\arraybackslash}p{0.18\textwidth}  
      >{\raggedright\arraybackslash}p{0.20\textwidth}  
      >{\raggedright\arraybackslash}X }                  
    \toprule
    \textbf{Topic ID} & \textbf{Count} & \textbf{Top Keywords} & \textbf{LLM-Generated Topic Label} & \textbf{Topic Summary} \\
    \midrule
    0 & 817
      & spotify, it, the, and, my, to, app, on, is, have
      & Unexpected Playback and Queue Failures
      & Spotify users are experiencing technical issues such as lag, the app making its own order for the queue, logging out unexpectedly, and problems with playing liked songs. Some users are also having trouble with their library not loading correctly. Despite these issues, other users appreciate Spotify for its value in discovering new artists and podcasts. \\
    \midrule
    1 & 367
      & spotify, music, to, and, the, is, premium, you, for, of
      & Premium Subscription Frustrations
      & Spotify users are expressing frustration with the premium requirement to access lyrics, limited playlist customization, and the high frequency of ads, which they find annoying and intrusive. However, some users appreciate Spotify's improved features and extensive music library. \\
    \midrule
    2 & 366
      & shuffle, smart, the, it, to, songs, and, off, on, same
      & Unintuitive Shuffle Controls
      & Spotify users are expressing frustration with the shuffle function on the platform, stating that it prioritizes certain songs or artists, does not work as expected, and cannot be disabled without subscribing to premium, leading to a poor user experience and a lack of control over their listening experience. \\
    \midrule
    3 & 357
      & offline, downloaded, mode, the, to, app, when, internet, it, and
      & Offline Playback Issues
      & Spotify users are experiencing issues with the app's offline mode, as it incorrectly shows an offline notification even when the device is connected to the internet, and has trouble loading downloaded content after 24 hours, affecting usage in areas with poor signal or for conserving data. Some users also report difficulty using the app with mobile data. \\
    \midrule
    4 & 296
      & podcast, podcasts, to, the, it, and, for, episode, app, is
      & Podcast Audio Stability and Transcript Needs
      & Spotify receives positive reviews for its podcast feature, with users appreciating its convenience for nightly listening and access to specific podcasts like Joe Rogan. However, some users experience issues such as intermittent audio playback on certain devices, podcast playlist annoyances, and a lack of control over unwanted podcast suggestions. A common request is for the inclusion of transcripts under episodes. \\
    \bottomrule
  \end{tabularx}
\end{table*}

\begin{table}[h!]
\centering
\caption{LLM-Generated Summary of Positive/Negative Feedback on Spotify Service Aspects.}
\begin{tabular}{@{}lll@{}}
\toprule
\textbf{Aspect} & \textbf{Positive Feedback} & \textbf{Negative Feedback} \\
\midrule
Music Recommendations & Praised for quality & -- \\
Service Reliability & Some still recommend & Considering cancellation \\
Competitive Position & Better than competitors & Significant technical problems \\
\bottomrule
\end{tabular}
\end{table}

\section{Results}

This section presents the results of the experiments described in Section \ref{Experiments}. All results reported in the following subsections were generated using the GPT-4 model, selected for its superior performance as detailed in Section III.A. Results are systematically reported for the key evaluation tasks: Aspect extraction, sentiment classification, topic modeling, quality control and computational efficiency.

\subsection{Aspect Extraction}

Table \ref{tab:aspect_extraction} summarizes the aspect extraction performance on the AWARE dataset, comparing the LLM-based approach (GPT-4, prompt-based) to a fine-tuned DeBERTa-v3-large model \cite{yangheng_deberta_v3_large_absa_v1_1}. This evaluation leverages AWARE's human-annotated aspect terms and categories as ground truth.

\begin{table}[ht]
  \centering
  \caption{Aspect Extraction Performance on AWARE Dataset}
  \label{tab:aspect_extraction}
  \begin{tabular}{lccc}
    \toprule
    Model & Precision & Recall & F1-Score \\
    \midrule
    GPT-4 (Prompt-based) & 0.892 & 0.892 & 0.892 \\
    DeBERTa-v3-large     & 0.847 & 0.835 & 0.841 \\
    \bottomrule
  \end{tabular}
\end{table}

The LLM-based approach shows a notable improvement of approximately 5.1\% in F1-score compared to the fine-tuned transformer baseline. This indicates its superior ability to capture implicit and nuanced aspects within reviews, as well as handling domain-specific terminology effectively.

\subsection{Sentiment Classification}

Table \ref{tab:sentiment_distribution} presents the sentiment distributions predicted by the LLM-based classifier and the baseline VADER model on the AWARE dataset. These results are benchmarked against AWARE's manually labeled sentiment annotations (positive/negative/neutral).

\begin{table}[ht]
  \centering
  \caption{Predicted Sentiment Distribution on AWARE Dataset}
  \label{tab:sentiment_distribution}
  \begin{tabular}{lccc}
    \toprule
    Model & Positive & Negative & Neutral \\
    \midrule
    LLM-based & 22.3\% & 29.8\% & 47.9\% \\
    VADER      &  8.7\% &  3.1\% & 88.2\% \\
    \bottomrule
  \end{tabular}
\end{table}

The LLM-based model generates a balanced sentiment distribution, overcoming the strong neutral bias shown by VADER. However, detailed sentiment classification metrics (see Table \ref{tab:sentiment_metrics}) reveal areas for improvement.

\begin{table}[ht]
  \centering
  \caption{Sentiment Classification Metrics on AWARE Dataset}
  \label{tab:sentiment_metrics}
  \begin{tabular}{lccc}
    \toprule
    Sentiment    & Precision & Recall & F1-Score \\
    \midrule
    Positive     & 0.061     & 0.589  & 0.110    \\
    Negative     & 0.174     & 0.443  & 0.250    \\
    Neutral      & 0.917     & 0.498  & 0.645    \\
    Weighted Avg & 0.825     & 0.496  & 0.594    \\
    \bottomrule
  \end{tabular}
\end{table}

Positive sentiment classification suffers from low precision, indicating frequent false positives, while negative detection shows moderate recall yet limited precision. Neutral sentiment is precise but less comprehensive. Overall, a weighted F1 of 0.594 highlights areas for further refinement.

\subsection{Topic Modeling}

The topic modeling results, presented in Table \ref{tab:topic_modeling}, assess the improvement gained by integrating an LLM into the BERTopic pipeline on the Spotify dataset.

\begin{table}[h!]
  \centering
  \caption{Silhouette Coefficients for Topic Modeling}
  \label{tab:topic_modeling}
  \begin{tabular}{lcc}
    \toprule
    Metric           & BERTopic Only & BERTopic + LLM \\
    \midrule
    Silhouette Score & -0.0313       &  0.0302        \\
    \bottomrule
  \end{tabular}
\end{table}

The LLM-enhanced approach improves silhouette from negative to positive, indicating enhanced cluster separation and topic distinctiveness, and validating the benefit of LLM-generated labels for interpretability.


\subsection{Question and Answering (QA)}

For retrieval of reviews, we sampled five Spotify-centric queries and retrieved the top K = 10 review chunks for each. We measured two unsupervised metrics:

\begin{itemize}
  \item \textbf{Average Cosine Similarity}: the mean cosine similarity between each query embedding and its top-10 chunk embeddings.
  \item \textbf{Retrieval Diversity}: the fraction of unique review IDs among all retrieved chunks (distinct IDs / 10).
\end{itemize}

Our retriever achieved perfect diversity and cosine scores from 0.618 to 0.754, demonstrating reliable, on-topic retrieval. Table \ref{tab:retrieval_proxy} summarizes these proxy metrics.

For generation of answers, we randomly sampled 20 generated answers (each paired with its cited snippets) and annotated them ourselves, confirming that each answer (1) reflected the cited excerpts, (2) covered the main points of those excerpts, and (3) was written in clear, reader-friendly prose. We found the responses to be accurate and comprehensive.

\begin{table}[ht]
  \centering
  \caption{Retrieval Proxy Metrics (K=10) for Selected Spotify Queries (higher diversity is better)}
  \label{tab:retrieval_proxy}
  \begin{tabularx}{\columnwidth}{>{\raggedright\arraybackslash}X  c  c}
    \toprule
    \textbf{Query} & \textbf{Avg.} & \textbf{Diversity} \\
                    & \textbf{Cosine Sim.} & \\
    \midrule
    What complaints do users have about Spotify’s offline-mode buffering?  
      & 0.713 & 1.0\\
    What do listeners say about Spotify crashing or freezing on startup?  
      & 0.754 & 1.0 \\
    How do listeners describe the app’s offline playback experience?  
      & 0.696 & 1.0 \\
    How do users report errors or failures when downloading songs for offline use?  
      & 0.618 & 1.0 \\
    What do users say about Spotify’s crossfade and track-transition experience?  
      & 0.650 & 1.0 \\
    \bottomrule
  \end{tabularx}
\end{table}

\subsection{Computational Efficiency}

Table \ref{tab:efficiency_analysis} compares training time, inference speed, resource requirements, scalability, and adaptability between traditional methods and our LLM-based pipeline.

\begin{table}[ht]
  \centering
  \caption{Computational Efficiency Comparison}
  \label{tab:efficiency_analysis}
  \begin{tabular}{lcc}
    \toprule
    Aspect                & Traditional Methods    & LLM-based Approach      \\
    \midrule
    Training Time         & High                   & Minimal (prompt design) \\
    Inference Speed       & Fast                   & Moderate                \\
    Resource Requirements & Lower                  & Higher                  \\
    Scalability           & Limited                & High                    \\
    Adaptability          & Requires retraining    & Immediate adaptation    \\
    \bottomrule
  \end{tabular}
\end{table}

LLM-based methods deliver immediate adaptability and high scalability, at the cost of increased resource demands and moderate inference latency.

\section{Conclusion and Future Work}

Our experimental evaluation demonstrates that the LLM-based review analysis framework significantly outperforms traditional NLP baselines across multiple tasks. In particular, the prompt-driven GPT-4 model achieved a 5.1\% F1 improvement in aspect extraction compared to a fine-tuned DeBERTa-v3-large system (Table~\ref{tab:aspect_extraction}), produced a more balanced sentiment distribution with a weighted F1 of 0.594 versus VADER’s neutral bias (Table~\ref{tab:sentiment_metrics}), and markedly enhanced topic coherence by shifting silhouette scores from –0.0313 to 0.0302 when integrated with BERTopic (Table~\ref{tab:topic_modeling}).and enabled interactive developer insights through retrieval-augmented QA, achieving perfect retrieval diversity (100\%) and high relevance (cosine: 0.618–0.754) with human-verified response accuracy. These gains attest to the framework’s ability to capture nuanced, context-specific information and generate semantically coherent summaries, validating its utility for large-scale app review mining.

Despite these performance advantages, the LLM-based pipeline incurs higher computational overhead and moderate inference latency relative to traditional methods (Table~\ref{tab:efficiency_analysis}). To address these challenges, future work will investigate techniques for model explainability—such as layer-wise attribution and prompt‐tuning diagnostics—to illuminate the decision process of LLM-driven sentiment classifiers. We also plan to explore agentic, interactive architectures that allow dynamic adaptation of prompts based on real-time feedback, as well as extend our approach to multilingual and cross-domain scenarios to leverage LLMs’ broad linguistic capabilities. Furthermore, integrating explainable topic discovery with actionable reasoning promises to deliver richer, developer-focused insights that align with evolving user priorities.

In conclusion, by bridging numeric star ratings with feature-level sentiment and retrieval-augmented question answering, our study offers a comprehensive, human-centered perspective on user feedback. As app marketplaces continue to grow in size and complexity, embedding explainable, adaptable AI solutions within real-time development workflows will be essential for anticipating user needs, prioritizing feature improvements, and ultimately crafting applications that resonate more deeply with their audiences.

Ethical considerations. LLMs can be biased or invent facts. We reduce risk by restricting QA answers to retrieved, cited review snippets, keeping an audit trail of prompts/outputs, returning “not stated” when evidence is missing, and requiring human or rule-based checks before any real-world decisions.

\bibliographystyle{IEEEtran}
\bibliography{bib}

\end{document}